\documentclass[letterpaper, 10pt,final,conference]{IEEEtran}

\usepackage{times}  % DO NOT CHANGE THIS
\usepackage{helvet}  % DO NOT CHANGE THIS
\usepackage{courier}  % DO NOT CHANGE THIS
\usepackage[hyphens]{url}  % DO NOT CHANGE THIS
\usepackage{graphicx} % DO NOT CHANGE THIS
\usepackage{caption} % DO NOT CHANGE THIS AND DO NOT ADD ANY OPTIONS TO IT
\usepackage[utf8]{inputenc}
\usepackage{comment}
\usepackage{cite}
\usepackage{booktabs}

\usepackage{graphicx}
\usepackage{subcaption}
\usepackage{amsmath}
\usepackage{verbatim}
\usepackage{amsfonts}
\usepackage{multirow}
\usepackage{changepage}
\usepackage{tabularx}
\usepackage{adjustbox}
\usepackage{mathtools}
\usepackage{amssymb}
\usepackage{soul}

\usepackage{color}
\usepackage{tikz}
\usepackage{flushend}
\usetikzlibrary{shapes.geometric, arrows}
\usepackage[linesnumbered,ruled]{algorithm2e}

\usepackage{soul,color}
\usepackage{graphicx}
\usepackage{fancyhdr}
\usepackage{comment}
\usepackage{epsfig}
\usepackage{bm}
\usepackage{float}
\usepackage{cuted}
\makeatletter
\def\algbackskip{\hskip-\ALG@thistlm}
\makeatother

\newcommand{\argmin}{\operatorname{arg min}}

\newcommand\oprocendsymbol{\hbox{$\square$}}
\newcommand\oprocend{\relax\ifmmode\else\unskip\hfill\fi\oprocendsymbol}

\newtheorem{remark}{Remark}

\begin{document}

\title{{ \LARGE \bf Towards Scalable \& Efficient Interaction-Aware Planning in Autonomous Vehicles using Knowledge Distillation}}

% \title{Unlocking Efficiency \& Scalability: Harnessing Knowledge Distillation for Neural Network-based Constrained Optimization
% }

\author{\IEEEauthorblockN{Piyush Gupta \hspace{1cm} David Isele \hspace{1cm} Sangjae Bae}
\IEEEauthorblockA{Honda Research Institute, San Jose, California, USA 95134\\
Email: \texttt{\{piyush{\_}gupta, disele, sbae\}@honda-ri.com }}}

\maketitle

% These inter-vehicle interactions enable complex maneuvers via intent sharing, yielding, negotiations, and cooperation. 

%Neural networks are increasingly used in engineering fields, but integrating them with traditional techniques often leads to complex constrained optimization problems. These problems are exacerbated by the non-convexity of neural networks,

\begin{abstract}
Real-world driving involves intricate interactions among vehicles navigating through dense traffic scenarios. Recent research focuses on enhancing the interaction awareness of autonomous vehicles to leverage these interactions in decision-making. These interaction-aware planners rely on neural-network-based prediction models to capture inter-vehicle interactions, aiming to integrate these predictions with traditional control techniques such as Model Predictive Control. However, this integration of deep learning-based models with traditional control paradigms often results in computationally demanding optimization problems, relying on heuristic methods. This study introduces a principled and efficient method for combining deep learning with constrained optimization, employing knowledge distillation to train smaller and more efficient networks, thereby mitigating complexity. We demonstrate that these refined networks maintain the problem-solving efficacy of larger models while significantly accelerating optimization. Specifically, in the domain of interaction-aware trajectory planning for autonomous vehicles, we illustrate that training a smaller prediction network using knowledge distillation speeds up optimization without sacrificing accuracy.
\end{abstract}

\begin{IEEEkeywords}
Interaction-aware planning, Knowledge distillation, Neural-network-based constrained optimization
\end{IEEEkeywords}

\section{Introduction}\label{Introduction}

{Navigating through dense traffic requires intricate inter-vehicle interactions, enabling complex maneuvers through intent sharing, yielding, negotiations, and cooperation. Autonomous Vehicles (AVs) sharing the road with human drivers must excel at comprehending and leveraging these interactions in their decision-making. Failing to do so may lead to excessive caution, disrupting traffic flow, and causing congestion. Moreover, an overly defensive AV could undermine passenger trust, resulting in a suboptimal experience. Figure~\ref{fig:interaction_unaware} illustrates an interactive scenario, highlighting the challenges an interaction-unaware AV may encounter, leading to extended waiting times. In contrast, Fig.~\ref{fig:interaction_aware} showcases an interaction-aware AV capable of predicting the consequences of its actions (such as indicating or nudging left) on other vehicles. This foresight enables the interaction-aware AV to make informed decisions.\\
\begin{figure}[]
    \centering
	\begin{subfigure}[b]{0.48\textwidth}
	    \centering
        \includegraphics[width=1\linewidth, height=1\linewidth, keepaspectratio]{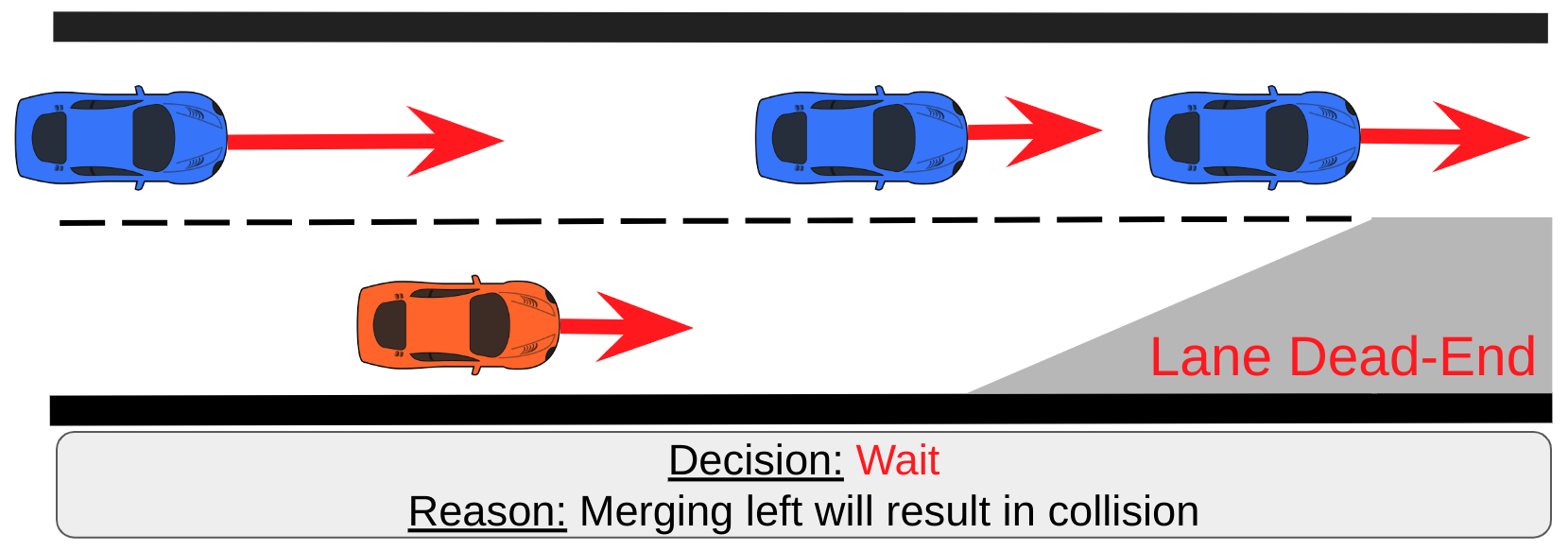}
        \caption{Interaction-unaware AV}
        \label{fig:interaction_unaware}
    \end{subfigure}
	\begin{subfigure}[b]{0.48\textwidth}
	    \centering
        \includegraphics[width=1\linewidth, height=1\linewidth, keepaspectratio]{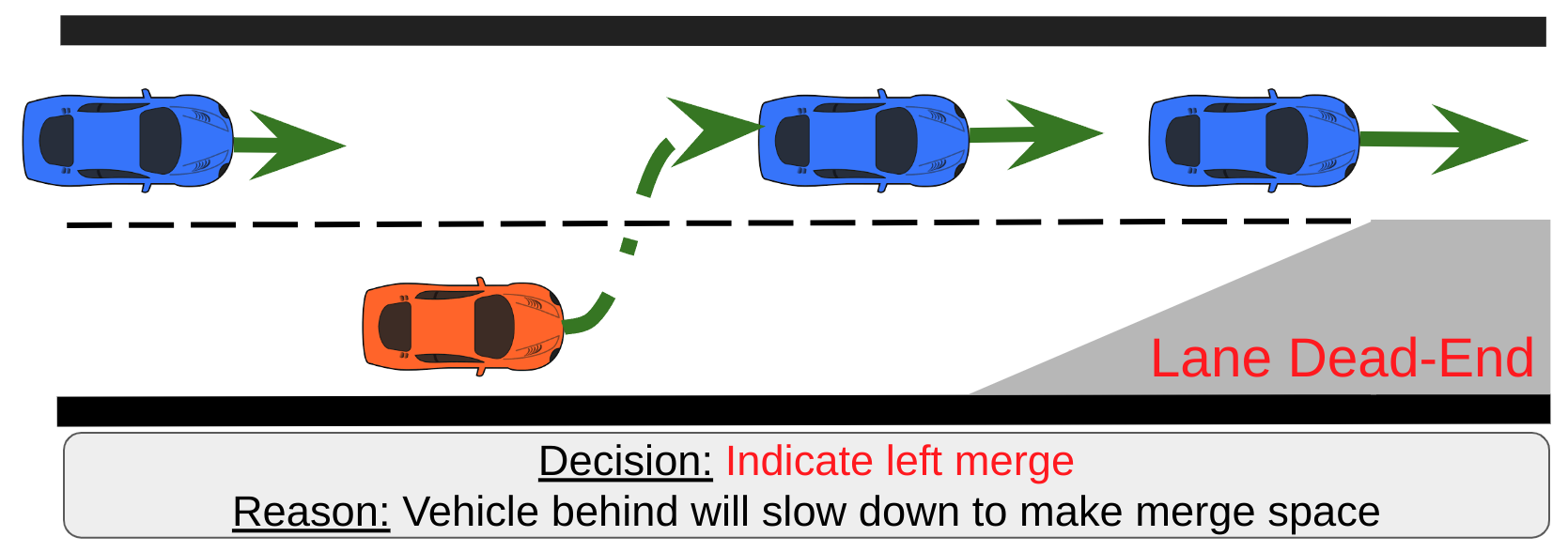}
        \caption{Interaction-aware AV}
        \label{fig:interaction_aware}
    \end{subfigure}
    \caption{\footnotesize{An illustrative scenario demonstrating the impact of interaction awareness. In (a), the interaction-unaware AV (orange) waits for an extended period, while in (b), the interaction-aware AV (orange) predicts that signaling or nudging left will prompt the blue vehicle behind the AV to decelerate, creating an opportune gap for a successful left merge.}}
    \label{fig:av_interaction}
\end{figure}
Recent research has emphasized improving the interaction awareness of AVs, with a central challenge being the effective capture of inter-vehicle interactions~\cite{zhou2023interaction, li2020behavior}. Deep learning methods, particularly in motion prediction~\cite{gupta2018social, sheng2022graph, martinez2017human}, have demonstrated notable success in capturing these interactions between agents. Despite the speed and accuracy of these neural networks, their use in safety-critical control tasks is often hindered by a lack of robust safety guarantees. Therefore, rather than relying solely on end-to-end deep learning systems for autonomous driving~\cite{coelho2022review, 9991836}, interaction-aware planners aim to integrate neural-network-based prediction models with traditional control techniques, such as Model Predictive Control (MPC)~\cite{kouvaritakis2016model}. This integration of optimization-based systems with data-driven methods leads to constrained optimization problems incorporating neural networks.

The neural-network-based constrained optimization in interaction-aware planning poses computational challenges. It leads to a closed-loop controller, using the MPC output (ego vehicle trajectory) as input for the prediction model and utilizing the prediction model's output to solve the MPC. Consequently, due to high computational complexity, prior approaches~\cite{9849019, sheng2022cooperation} have relied on heuristic methods for trajectory generation, resulting in sub-optimal solutions. In contrast, \cite{10160890} proposes an optimization-based planner that has been demonstrated to converge to locally optimal solutions, presenting a desirable property for the planner. However, the computational complexity of the prediction model in \cite{10160890} constrains the optimization speed, limiting scalability and rendering it as an offline method. Our research addresses this challenge by seeking to improve the computational efficiency of the optimization procedure and enhance the scalability of the approach.

Knowledge distillation~\cite{gou2021knowledge} offers an effective methodology for training a smaller `student' neural network, drawing upon the knowledge from a larger, pre-trained `teacher' network. This technique facilitates the distillation of the teacher network's knowledge, enabling the consolidation of this information within a more compact neural network. Although previous studies have evaluated the efficacy~\cite{cho2019efficacy} and similarity-preservation capability~\cite{tung2019similarity} of knowledge distillation, its influence on neural-network-based optimization problems remains under-explored in the literature. These neural-network-based optimization problems are encountered in robotics systems that utilize neural networks for upstream tasks and traditional closed-loop control methods for downstream tasks. Hence, this work aims to probe the impact of implementing a student network, trained via knowledge distillation, on the efficiency and precision of solving neural-network-based, constrained non-convex optimization problems. Our research is motivated by the real-world hurdles encountered when deploying systems that seek to integrate data-driven methods with traditional control and optimization techniques. 
% \di{this is good, maybe just remove the previous 2 paragraphs. but keep the citations}

% To examine the effect of knowledge distillation on constrained optimization, we focus on the context of interaction-aware trajectory planning in autonomous driving, as outlined by~\cite{10160890}. This planning framework requires the integration of a neural-network-based interactive prediction model with conventional Model Predictive Control (MPC). Previous approaches~\cite{sheng2022cooperation, 9849019} have utilized heuristic methods for trajectory generation within this framework, resulting in sub-optimal solutions. In contrast, \cite{10160890} propose an optimization-based planner that converges to locally optimal solutions. However, the computational complexity of the prediction model limits the speed of optimization, hindering scalability. Our research addresses this challenge by seeking to improve the computational efficiency of the optimization procedure and enhance the approach's scalability.

While our primary focus is on interaction-aware trajectory planning for AVs, our work has broad applications in domains involving closed-loop control systems with neural networks, leading to neural-network-based optimization. For instance, a robot arm can utilize a neural network to adapt its control inputs for grasping objects of various shapes and sizes based on visual or tactile feedback. In aviation, neural networks can enhance aircraft stability and control during flight by adjusting factors like turbulence response and aircraft state. Our results, particularly in knowledge distillation for neural-network-based optimization, offer viable solutions in three key scenarios: (i) tapping into the knowledge embedded in large foundational models~\cite{yuan2023power} for specific tasks, (ii) using the network as a black-box model in optimization, and (iii) implementing system-specific architectural changes, such as requiring twice-differentiable activation for achieving convergence (see~\cite{10160890}). Importantly, our contribution facilitates the utilization of the rigorous convergence properties demonstrated in \cite{10160890} in real-time applications.}

% \blue{
% Neural networks pose various challenges in constrained optimization. Retraining pre-trained networks via transfer learning~\cite{torrey2010transfer} for specific applications can burden computation due to their complex model structures.  For instance, versatile foundation models~\cite{bommasani2021opportunities} may be used for a specific task, which could be adequately handled by a substantially smaller model.  Moreover, the absence of explicit mathematical architecture for these pre-trained networks poses significant challenges when they are employed as black-box models within optimization scenarios. Additionally, certain system designs may require unique architectural modifications not met by standard pre-trained models, necessitating custom neural network designs (see~\cite{10160890} for convergence requirements). Despite these hurdles, it remains crucial to draw from the knowledge encapsulated within these trained networks to design application-specific networks.
% }

This work presents two key contributions. Firstly, we employ knowledge distillation to train a compact interactive prediction network, which efficiently predicts the impact of the ego vehicle's trajectory on other vehicles in a single inference. This approach streamlines the prediction process and enhances efficiency of the optimization. Secondly, we demonstrate a significant fivefold improvement in computation time by utilizing the student network to solve the neural-network-based constrained optimization problem without any significant loss in accuracy. This outcome underscores the effectiveness of our approach in accelerating the optimization process while preserving accuracy. 

The remaining paper is structured as follows. Section~\ref{sec:background} provides background and formulation for the neural-network-based constrained optimization, with emphasis on its application to interaction-aware trajectory planning for AVs. Section~\ref{sec:knowledge_distillation} introduces our proposed solution to enhance the computational efficiency of the interaction-aware planner. We describe the design and training of a smaller interactive prediction model, referred to as the student network, which learns from a larger prediction model, the teacher network, through knowledge distillation. In Section~\ref{sec:results}, we present the results obtained from comparing the teacher and student networks in the motion planner. Finally, we conclude in Section~\ref{sec:conclusions}. 

\section{Background and Formulation}\label{sec:background}

\begin{figure}[ht]
    \centering
    \includegraphics[width=1\linewidth, height=\linewidth, keepaspectratio]{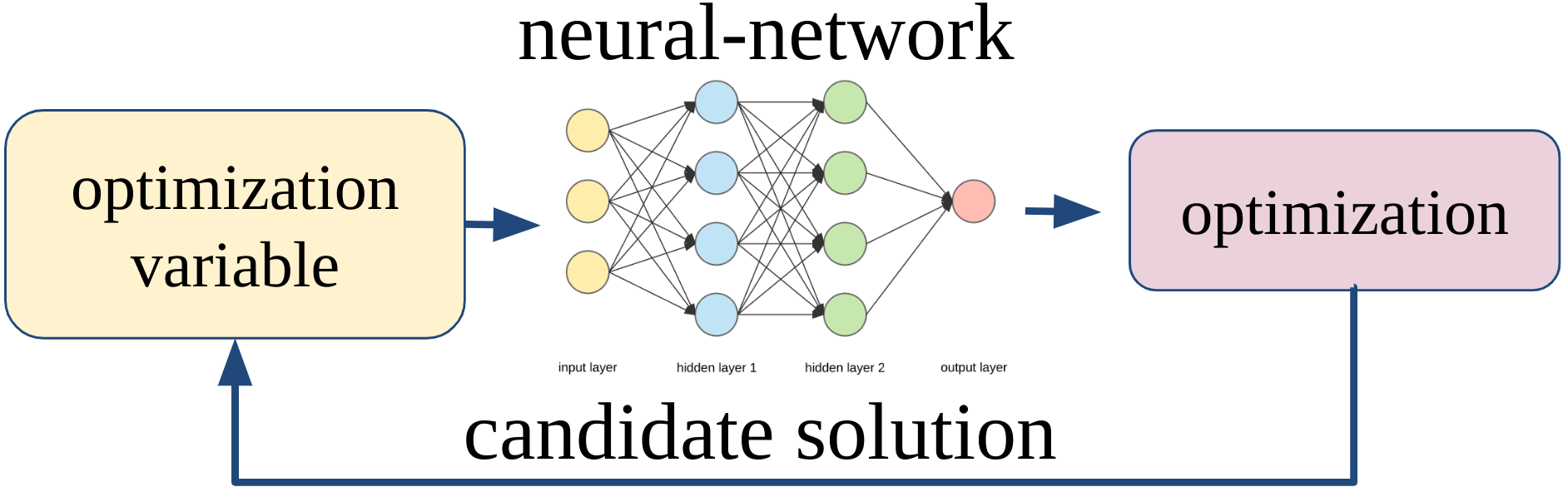}
    \caption{\footnotesize{{Neural-network-based optimization integrates a neural network with closed-loop control, utilizing the neural network's output for optimization and feeding the optimization output back into the neural network.}}} 
    \label{fig:closed_loop}
\end{figure} 

The widespread use of data-driven approaches is supported by the high efficiency and accuracy of neural networks. Their representation capability effectively approximates highly complex systems, leading to active analysis of their integration into conventional optimization-based methods~\cite{kittisupakorn2009neural, sadigh2016planning, bae2020cooperation}. This integration gives rise to neural-network-based optimization problems (see Fig.~\ref{fig:closed_loop}) where the neural network's output becomes part of the objective or constraint formulation, and the output of the optimization serves as an input to the neural network. However, solving these optimization problems efficiently can be challenging due to the inherent non-convexity and complexity of neural networks.

Consider a general non-linear program (NLP)~\cite{bertsekas1997nonlinear} that incorporates a neural-network-based formulation for functions $f$, $g$, and $h$:
\begin{align}
&\min_{x} f(x, \phi(x ; w)) \label{eq:NLP_1} \\
\text{subject to } \ \ \  & g(x, \phi(x ; w)) = 0 \label{eq:NLP_2}\\
& h(x, \phi(x ; w)) \ge 0, \label{eq:NLP_3}
\end{align}
where, $x \in \mathbb{R}^m$ denotes the vector of optimization variables, $\phi(x; w): x \xrightarrow{} \mathbb{R}^n$ represents a pre-trained neural network with weights $w$, and $m$ and $n$ are arbitrary positive integers. In various real-world applications, the neural network $\phi(x;w)$ often appears exclusively in one of the functions: $f$, $g$, or $h$~\cite{bae2020cooperation}. Moreover, when $\phi(x;w)$ is absent (or replaced with a constant), the optimization problem often represents a simpler linear program (LP) or a quadratic program (QP). Nevertheless, the non-convex nature and intricate structure of $\phi(x;w)$ pose significant challenges in solving the optimization problem in real-time, irrespective of the convexity and simplicity of the functions $f$, $g$, and $h$.

Various optimal control solvers are available for Nonlinear Programming (NLP) problems like those defined in~\eqref{eq:NLP_1}-\eqref{eq:NLP_3}. However, finding a generic solver that works well in all cases is unrealistic. One popular choice is IPOPT~\cite{biegler2009large}, a general NLP solver that uses an interior-point line-search filter method. These solvers typically rely on gradient-based optimization methods that use first and second derivative (Hessian) information to find the optimal solution. When Hessian is unavailable, quasi-Newton methods~\cite{schoenberg2001optimization}, such as the Broyden–Fletcher–Goldfarb–Shanno (BFGS) method~\cite{liu1989limited}, are often used to approximate the Hessian.

When dealing with NLP problems involving neural networks, such as those in~\eqref{eq:NLP_1}-\eqref{eq:NLP_3}, computing the gradients with respect to the optimization variables $x$ requires back-propagation of the gradients through the neural network $\phi(x;w)$ to calculate $\frac{\partial \phi(x;w)}{\partial x}$. Additionally, running back-propagation and forward inference multiple times is often necessary to solve the optimization. However, performing these operations can be computationally expensive and infeasible for many real-time applications.

\begin{remark}
It should be noted that the $\phi(x;w)$ represents a pre-trained neural network with fixed parameters $w$. The back-propagation of gradients mentioned earlier is distinct from the back-propagation utilized during network training, where the derivatives of the loss with respect to the weights and biases are computed. In this case, back-propagation refers to the computation of the derivatives of the network's outputs with respect to its inputs $x$, which also serve as the optimization variables.
\end{remark}

This study focuses on scaling the interaction-aware trajectory planner for autonomous vehicles, as outlined in the work by~\cite{10160890}. The proposed approach combines a neural-network-based interactive prediction model with an MPC-based trajectory planner. The integration of these components leads to a simplified version of the general neural-network-based NLP, as described in~\eqref{eq:NLP_1}-\eqref{eq:NLP_3}, which poses challenges for real-time solution.

To address this challenge, we investigate the impact of employing smaller interactive prediction models in the optimization process. These models are designed and trained using knowledge distillation techniques using the knowledge encapsulated in the larger prediction model. By utilizing these smaller models, the goal is to enhance real-time performance while securing provable optimality provided by ~\cite{10160890}. In the interest of completeness, we now summarize the interaction-aware trajectory planner. 

\subsection{Interaction-Aware Trajectory Planning}\label{subsec:interaction_aware_trajectory_planning}

Autonomous Vehicles (AVs) must operate in shared spaces alongside human drivers, resulting in interactive tasks where the actions of AVs impact nearby vehicles and vice versa~\cite{ ulbrich2013probabilistic,sadigh2016planning,isele2019interactive,tian2021anytime}. To leverage these interactions in AV's motion planning and decision-making, a neural-network-based interactive prediction model is integrated with MPC, resulting in a neural-network-based NLP. %The interactive prediction model is based on Social-GAN~\cite{gupta2018social}.

% \blue{the former algorithm requires iterative inferences as the control input is updated iteratively. Instead, we train a network with a sequence of control solutions as a state, thus avoiding iterative process but having a one-shot inference over the planning horizon. }

The optimization problem for receding horizon control with a planning horizon $t_{plan}$ and a set of surrounding vehicles $\mathcal{V}$ can be formulated as follows:
\begin{align}
    \min_{\boldsymbol{\Delta},\boldsymbol{\alpha}, \boldsymbol{Z}}\;\; J=& \Phi_1(\boldsymbol{\Delta}) + \Phi_2(\boldsymbol{\alpha})+ \Phi_3(\boldsymbol{Z}),\label{eq:compact_obj}\\
    \text{subject to  } \;\;& 
    F(\boldsymbol{\Delta}, \boldsymbol{\alpha}, \boldsymbol{Z})  =0, \label{eq:equality_constraint}\\
    & b_i(\boldsymbol{Z}, \phi(\boldsymbol{Z}; w))>0, \ i \in \mathcal{V}. \label{eq:inequality_constraint}
\end{align}
In this formulation, the optimization variables $\boldsymbol{\Delta} \in \mathbb{R}^{t_{plan}}$, $\boldsymbol{\alpha}\in \mathbb{R}^{t_{plan}}$, and $\boldsymbol{Z}\in \mathbb{R}^{4t_{plan}}$ represent the trajectories of steering, acceleration, and vehicle state (xy-coordinates, heading angle, speed) respectively, over the planning horizon $t_{plan}$. The functions $\Phi_i(\cdot)$ (where $i \in {1,2,3}$) are quadratic functions, $F$ represents the linearized bicycle kinematics (system dynamics constraints), and $b_i(\cdot)$ denotes the safety constraints for the AV with respect to the surrounding vehicle $i  \in \mathcal{V}$. The interactive neural-network-based prediction model $\phi(\cdot)$ is integrated into the safety constraints $b_i(\cdot)$. Importantly, due to the interactions, the output of the neural network $\phi(\cdot)$ depends on the ego vehicle's states (see Section~\ref{subsec:Interactive_Prediction_Model}), which are also considered as optimization variables. Interested readers are referred to~\cite{10160890} for further details in formulations.

The formulated optimization problem is challenging to solve due to the presence of an integrated neural network and a large number of optimization variables. However,~\cite{10160890} proposed a solution approach that decouples the optimization variables $\boldsymbol{\Delta}$, $\boldsymbol{\alpha}$, and $\boldsymbol{Z}$, allowing the use of the alternative direction method of multipliers (ADMM)~\cite{boyd2011distributed}. By introducing the augmented Lagrangian $\mathcal{L}_\rho$ defined as follows:
\begin{multline}
\mathcal{L}_\rho(\boldsymbol{\Delta},\boldsymbol{\alpha}, \boldsymbol{Z}) = \Phi_1(\boldsymbol{\Delta}) + \Phi_2(\boldsymbol{\alpha}) + \Phi_3(\boldsymbol{Z}) \\
+ \mu^\top F(\boldsymbol{\Delta},\boldsymbol{\alpha}, \boldsymbol{Z}) + \left(\frac{\rho}{2}\right)\lVert F(\boldsymbol{\Delta},\boldsymbol{\alpha}, \boldsymbol{Z}) \rVert^2,
\end{multline}
where $\rho > 0$ is the ADMM Lagrangian parameter and $\mu$ is the dual variable associated with the constraint~\eqref{eq:equality_constraint}, the original optimization problem can be solved through an iterative process involving three sub-optimizations (see Algorithm~\ref{alg:mpc_admm}). 
\begin{algorithm}[ht]
\small
   \SetKwInOut{Init}{Init}
        \While{convergence criterion is not met}{
            $\boldsymbol{\hat{\Delta}} \leftarrow \argmin_{\boldsymbol{\Delta}} \mathcal{L}_\rho(\boldsymbol{\Delta},\boldsymbol{\hat{\alpha}}, \boldsymbol{\hat{Z}})$, \ \text{s.t.} \ $\boldsymbol{\Delta} \in [\delta_{\min}, \delta_{\max}]$ \\
            $\boldsymbol{\hat{\alpha}} \leftarrow \argmin_{\boldsymbol{\alpha}} \mathcal{L}_\rho(\boldsymbol{\hat{\Delta}},\boldsymbol{\alpha}, \boldsymbol{\hat{Z}})$, \ \text{s.t.} \ $\boldsymbol{\alpha} \in [\alpha_{\min}, \alpha_{\max}]$ \\
            $\hat{\boldsymbol{Z}} \leftarrow \argmin_{\boldsymbol{Z}} \mathcal{L}_\rho(\boldsymbol{\hat{\Delta}},\boldsymbol{\hat{\alpha}}, \boldsymbol{Z})$, \ \text{s.t.} \ $b_i(\boldsymbol{Z}, \phi(\boldsymbol{Z}; w))>0, \ i \in \mathcal{V}$  \\
            $\hat{\mu}\leftarrow \hat{\mu}+\rho(F(\boldsymbol{\hat{\Delta}},\boldsymbol{\hat{\alpha}}, \boldsymbol{\hat{Z}}))$\\
        }
    \caption{MPC with ADMM}\label{alg:mpc_admm}
\end{algorithm}

In Algorithm~\ref{alg:mpc_admm}, it is important to highlight that the updates for steering ($\boldsymbol{\hat{\Delta}}$) and acceleration ($\boldsymbol{\hat{\alpha}}$) are quadratic problems that can be efficiently solved. However, the state update ($\boldsymbol{\hat{Z}}$) involves non-convex neural network predictions, leading to higher computational complexity. This update becomes a bottleneck in terms of computational efficiency, and our goal is to expedite this optimization process by using knowledge distillation.

{
To address the high computational complexity of the $\boldsymbol{\hat{Z}}$ update, we adopted a projected gradient descent method that solely relies on the gradient of the objective function, eliminating the need for the gradient of the safety constraint. This approach avoids the computationally expensive step of back-propagating gradients through the neural network, as the neural network only appears in the safety constraints. Although this improves the computational complexity of the optimization process, achieving real-time performance is still challenging due to the requirement of multiple forward neural-network inferences for the optimization.

During the $\boldsymbol{\hat{Z}}$ optimization, the gradient descent updates the optimization variable $\boldsymbol{Z}$ (candidate solution) by moving it in the direction of the negative gradient of the objective function. After each iteration of the gradient descent, it is crucial to assess the candidate solution's adherence to safety constraints by executing the interactive prediction model with the candidate solution.

To compute these interactive predictions, the former method in~\cite{10160890} adopts a recursive approach that iteratively uses the prediction model $\phi(\tau)$ with an iterative update of the ego vehicle's trajectory using the candidate solution. This process results in the teacher model which is presented in Section~\ref{subsec:teacher_network}. In contrast, we employ knowledge distillation to train a network (student network in Section~\ref{subsec:student_network}) that takes the ego vehicle's candidate solution as input, avoiding the iterative process and enabling a one-shot inference over the planning horizon. This allows us to effectively mitigate computational complexity, facilitating real-time performance. 
}

\subsection{Interactive Prediction model}\label{subsec:Interactive_Prediction_Model}
The interactive prediction model $\phi(\tau)$
% is based on Social-GAN \cite{gupta2018social} \bae{we should not limit our work to this very specific neural network. Please add a remark or a statement that any types of neural networks satisfying the conditions (in the ICRA paper) can be leveraged. Plus, we would want to put a short motivation behind using SGAN as a particular example.} and 
predicts the future trajectories of the ego vehicle and its surrounding vehicles. It considers the past $t_{obs}$ time-steps of their trajectories and generates predictions for the future $t_{pred}$ time-steps. At any given time $\tau$, when $t_{pred}=1$, the interactive prediction model $\phi(\tau)$ can be represented as:
\begin{align}\label{eq:SGAN}
    \phi(\tau) : 
    &\begin{bmatrix}
    (x(\tau),y(\tau))&\cdots&(x_{N}(\tau),y_{N}(\tau))\\
    \vdots&\vdots&\vdots\\
    \begin{matrix}(x(\tau-t_{obs}+1),\\ \ y(\tau-t_{obs}+1))\end{matrix} &\cdots&\begin{matrix}(x_{N}(\tau-t_{obs}+1),\\ \ y_{N}(\tau-t_{obs}+1))\end{matrix}
    \end{bmatrix}\nonumber\\
    & \hspace{4cm} \downarrow \nonumber \\
    &\!\!\!\!\!\! \begin{bmatrix}
    (\hat{x}(\tau+1),\hat{y}(\tau+1))& \cdots&(\hat{x}_{N}(\tau+1),\hat{y}_{N}(\tau+1))
    \end{bmatrix},
\end{align}
where the first column denotes the positions of the ego vehicle, followed by the positions of $N$ surrounding vehicles. The model $\phi(\tau)$ takes into account the past positions from $\tau-t_{obs}+1$ to $\tau$ and predicts the future positions of the ego vehicle ($\hat{x}, \hat{y}$) and the surrounding vehicles  ($\hat{x}_i, \hat{y}_i, i \in \mathcal{V}$) at time $\tau+1$. It is worth noting that although the model has the capability to generate predictions for $t_{pred}>1$, such predictions would lack interactivity. Therefore, the planned trajectory of the ego vehicle would not influence the predictions of the surrounding vehicles.

To achieve interactive predictions over the planning horizon $t_{plan}$, a recursive approach is employed. The prediction model $\phi(\tau)$ is used repeatedly for $t_{plan}$ time-steps with $t_{pred} = 1$. In this process, the ego vehicle positions in the output of the prediction model are substituted with the candidate solution trajectory in the MPC optimization. This allows for the interactive influence of the ego vehicle's planned trajectory on the predictions of the surrounding vehicles throughout the planning horizon.

{\begin{remark}
It is important to note that $\phi(\tau)$ can be any pre-trained prediction model of the form \eqref{eq:SGAN}. Some examples of such models include the social generative adversarial network (Social-GAN)~\cite{gupta2018social} and the graph-based spatial-temporal convolutional network (GSTCN)~\cite{sheng2022graph}. For the purpose of this study, we utilize the Social-GAN network, which has been widely employed in the literature as an interactive prediction model~\cite{bae2020cooperation, 10160890}.
\end{remark}}

In the next section, we discuss knowledge distillation as a solution to enhance the computational efficiency of the interaction-aware trajectory planner.

\section{Knowledge Distillation}\label{sec:knowledge_distillation}

Deep neural networks often face limitations due to their large size, which restricts their usage on constrained compute systems. However, knowledge distillation offers an effective solution by training smaller student models from larger teacher models. This approach compresses and accelerates the models, making them suitable for devices with limited resources, such as mobile phones and embedded devices. In this work, we utilize knowledge distillation to enhance the computational complexity of the interaction-aware trajectory planner, which combines a neural-network-based prediction model with MPC.

\subsection{Teacher prediction model}\label{subsec:teacher_network}

\begin{figure}[ht]
    \centering
    \includegraphics[width=1\linewidth, height=\linewidth, keepaspectratio]{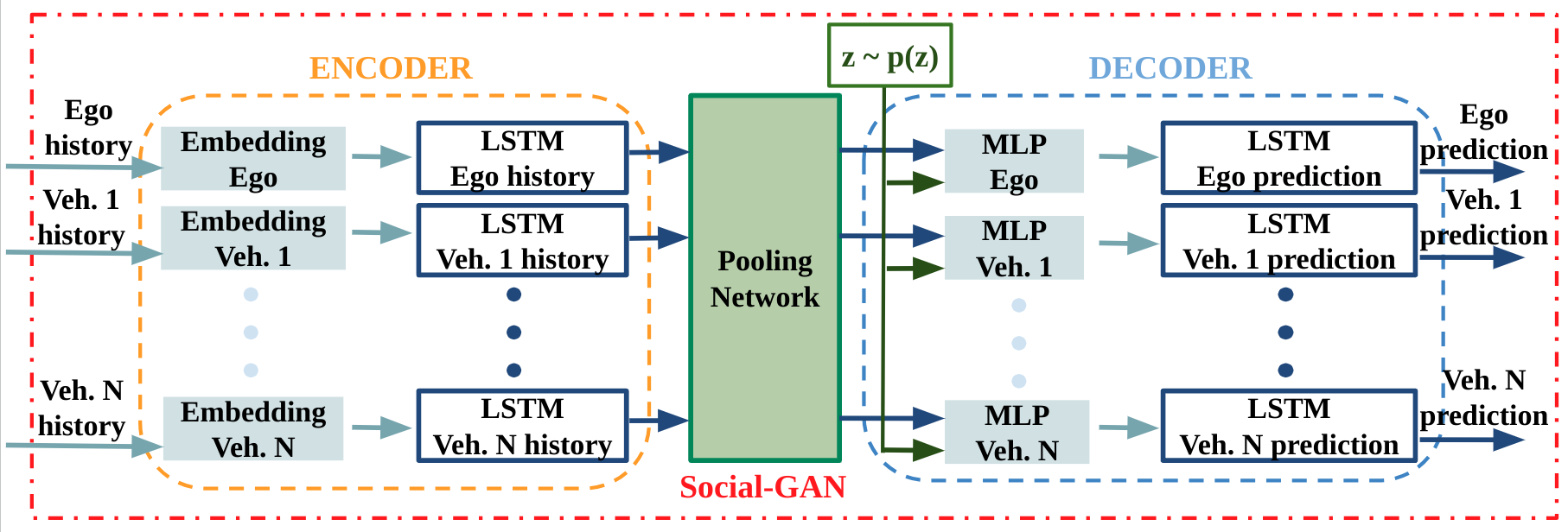}
    \caption{\footnotesize{Overview of the Social-GAN's RNN based Encoder-Decoder generator for a one time-step prediction ($t_{pred} = 1$).}} 
    \label{fig:sgan_generator}
\end{figure} 

\begin{figure*}[ht]
    \centering
	\begin{subfigure}[b]{0.46\textwidth}
	    \centering
        \includegraphics[width=1\linewidth, height=1\linewidth, keepaspectratio]{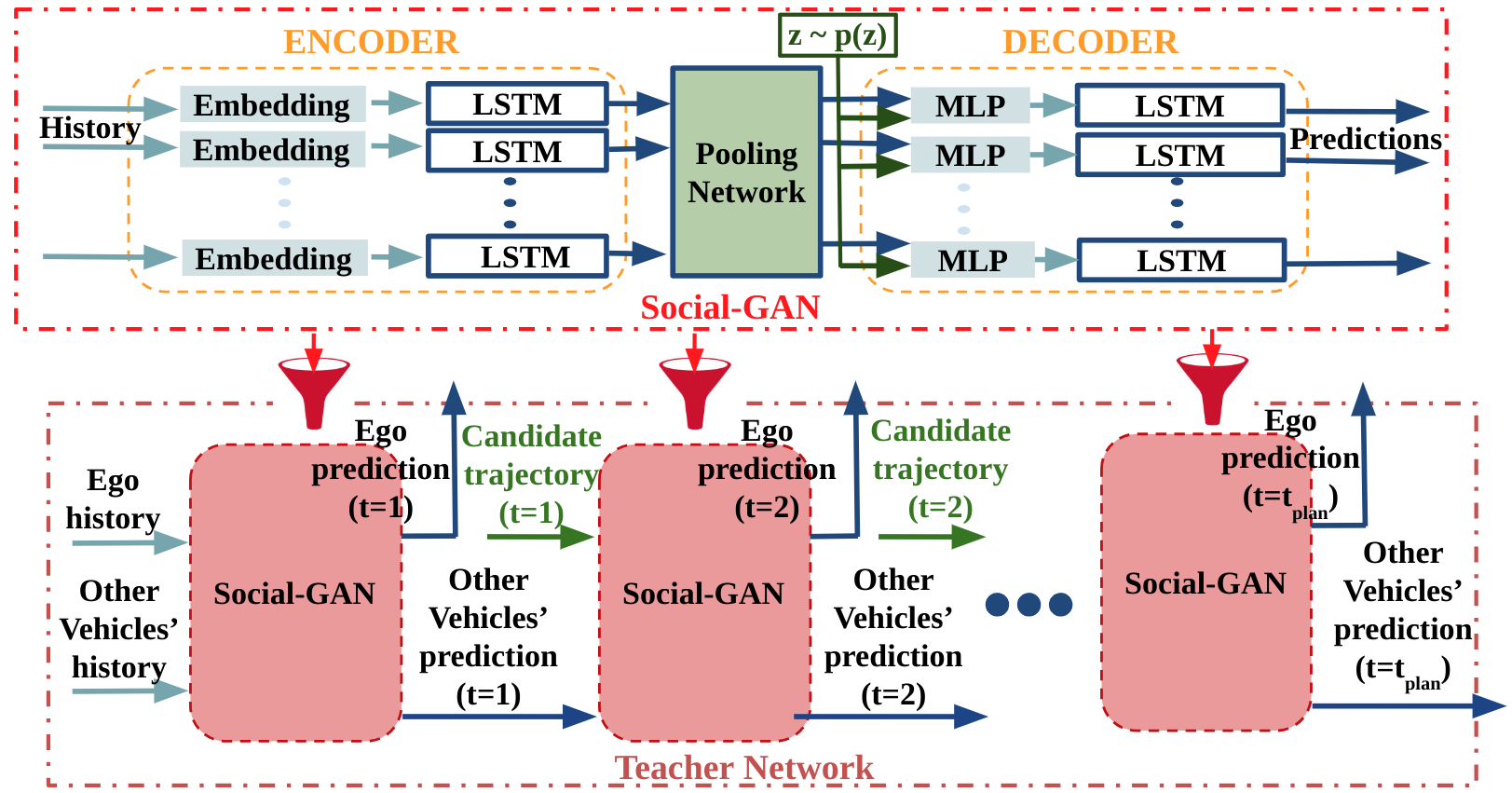}
        \caption{Teacher Network Generator}
        \label{fig:teacher_network}
    \end{subfigure}
	\begin{subfigure}[b]{0.53\textwidth}
	    \centering
        \includegraphics[width=1\linewidth, height=1\linewidth, keepaspectratio]{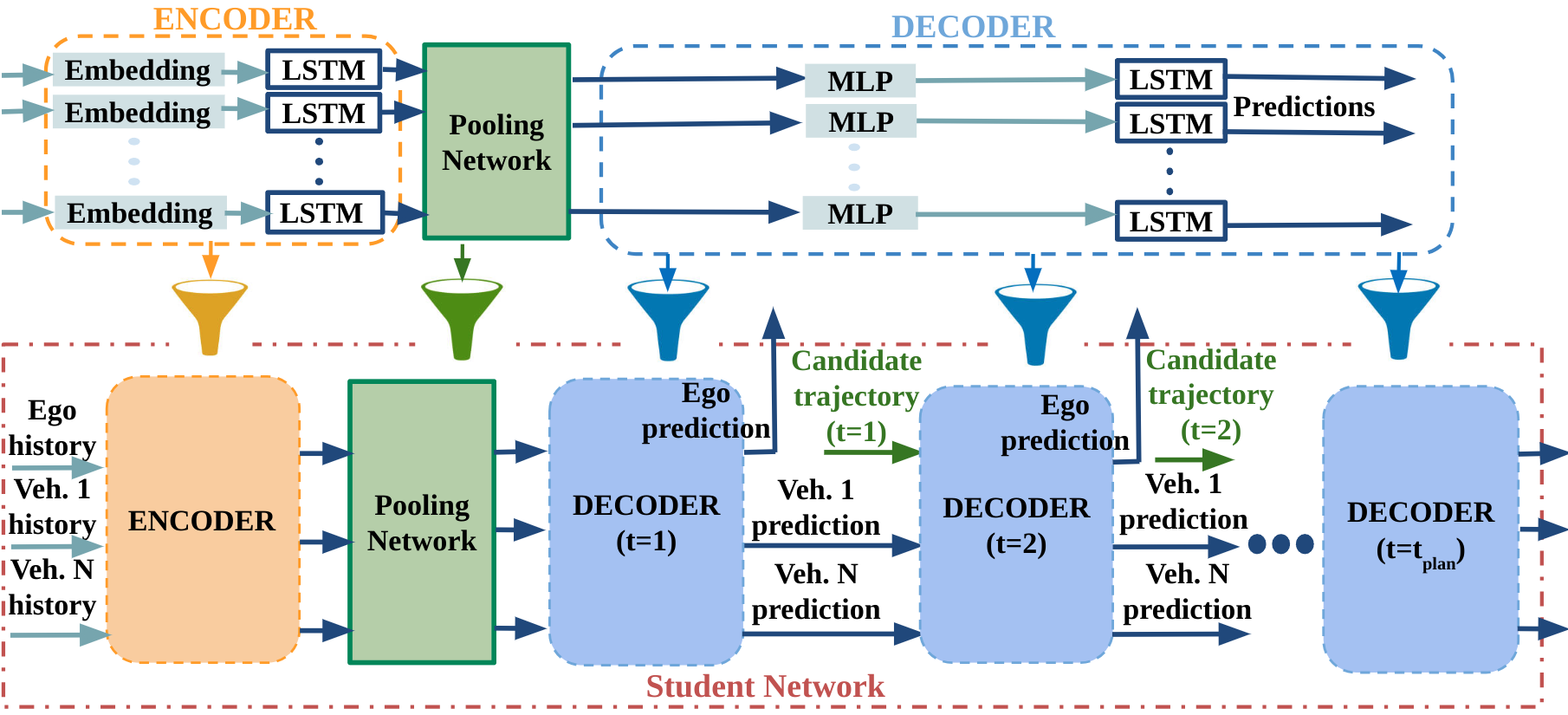}
        \caption{Student Network Generator}
        \label{fig:student_generator}
    \end{subfigure}
    \caption{\footnotesize{ Overview of the interactive trajectory generator for the (a) teacher network and the (b) student network. (a) The teacher network utilizes Social-GAN's generator multiple times to generate interactive trajectories for $t_{plan}$ time steps. (b) The student network produces the complete interactive trajectory by taking the ego candidate trajectory as input and running the decoder for $t_{plan}$ time steps. }}
    \label{fig:knowledge_distillation}
\end{figure*}

{We start with the interaction-aware trajectory planner proposed by~\cite{10160890} that employs a deep learning based prediction framework such as the Social-GAN based prediction model~\cite{gupta2018social}.}
% The original interaction-aware trajectory planner \di{consider change: We start with the interaction-aware trajectory planner proposed by...} ~\cite{10160890} employs \di{a deep learning based prediction framework, for example} the Social-GAN based prediction model (see~\eqref{eq:SGAN}) \di{cite SGAN}. 
The Social-GAN prediction model consists of an RNN-based Encoder-Decoder generator and an RNN-based encoder discriminator. Fig.~\ref{fig:sgan_generator} shows an overview of the Social-GAN's generator model for a one time-step prediction ($t_{pred} = 1$). A latent variable $z$ is drawn from a standard normal distribution $z \sim \mathcal{N}(0,1)$, enabling it to capture multiple prediction modalities and generate diverse samples. During each step of the MPC optimization, interactive predictions for surrounding vehicles are crucial over a planning horizon $t_{plan}$. To be interactive, these predictions must consider the influence of the ego-vehicle's trajectory, a subset of the optimization variables, and hence, must be derived from both the historical vehicle data and the ego vehicle's candidate trajectory. Consequently, the generator in Fig.~\ref{fig:sgan_generator} is repeatedly used for $t_{plan}$ time-steps, replacing the ego-vehicle's prediction with its candidate trajectory at each step. Fig.~\ref{fig:teacher_network} shows the interactive prediction model (teacher network) that uses Social-GAN's generator for $t_{plan}$ time-steps to produce interactive predictions.

% \begin{figure}
%     \centering
%     \includegraphics[width=0.95\linewidth, height=\linewidth, keepaspectratio]{images/teacher_network.png}
%     \caption{\footnotesize{Teacher network runs Social-GAN's generator multiple times to produce the interactive trajectories for $t_{plan}$ time steps.}} 
%     \label{fig:teacher_network}
% \end{figure} 

\textbf{Teacher Network Training Dataset:} To train the teacher model, scenarios from the Next Generation Simulation (NGSIM) dataset obtained from the Interstate 80 Emeryville, CA freeway were utilized\footnote{\url{https://ops.fhwa.dot.gov/trafficanalysistools/ngsim.htm}}. The scenarios were categorized into interactive, involving lane-change and merging scenarios, and non-interactive, consisting of lane-keeping scenarios. Noisy and unrealistic data were filtered out to ensure high-quality data. After filtering, a dataset of $68,220$ interactive examples and $24,275$ non-interactive examples was collected. Each example included joint-vehicle trajectories for anywhere from $1$ to $47$ vehicles including the ego vehicle. The dataset was further refined to remove examples with missing data, resulting in a final dataset of $91,387$ examples. For training and evaluation, an 80/20 random split was performed, with the training data further divided into a 90/10 ratio for training and cross-validation. Consequently, the training, cross-validation, and test sets consisted of approximately $66$K, $7.5$K, and $18$K examples, respectively.

%$65,798$, $7,310$, and $18,277$ examples, respectively.

\textbf{Teacher Network Training:} LSTM is employed as the chosen RNN for both the encoder and decoder components of the models. The hidden state dimension is set to $32$ for the encoder and decoder in the generator, and $48$ for the encoder in the discriminator. The MLP component consists of a single hidden layer with a dimension of $64$. Input coordinates from the historical data of each vehicle for $t_{obs}=8$ time-steps are embedded into $16$-dimensional vectors. The models were trained for $200$ epochs with a batch size of $64$ using the ADAM optimizer~\cite{kingma2014adam} with an initial learning rate of $0.0001$.

We employ two different training mechanisms to train multiple prediction models for the teacher network: regression and adversarial training (GAN) with a discriminator network. Following the approach proposed in~\cite{gupta2018social}, we generate $k$ possible output predictions ($k=20$ for trained models) for each scenario by randomly sampling the latent variable $z$ from $\mathcal{N}(0,1)$. These output predictions are used in the variety loss, which selects the ``best" prediction in terms of the L2 sense. This encourages the network to generate diverse samples and has been shown to significantly improve accuracy~\cite{gupta2018social}. The loss function for the generator network is defined as:
\begin{align}\label{eq:loss}
\mathcal{L}_{G} &= k_{var}\mathcal{L}_{var} + \mathbb{I}_{adv}\mathcal{L}_{adv} \\
& = \frac{k_{var}}{\lVert V \rVert} \min_{k} \sum_{i \in V}\lVert Y_i - \hat{Y}_i^{(k)}\rVert_2 + \mathbb{I}_{adv} \mathcal{L}_{adv},
\end{align}
where the first component represents the variety loss, while the second component represents the standard adversarial loss given by $\mathbb{E}_{z \sim p(z)}[log(1 - D(G(z)))]$ for the generator $G$ and discriminator $D$. Here, $Y_i$ and $\hat{Y}_i$ represent the ground truth trajectory and predicted trajectory for vehicle $i$, respectively. $V$ denotes the set of all vehicles in the training example, and $\lVert V \rVert$ represents the total number of vehicles including the ego vehicle. The hyper-parameter $k_{var}$ controls the impact of the variety loss, and $\mathbb{I}_{adv}$ is the indicator function. It is set to $1$ during adversarial training and $0$ when training as a regression model using only the variety loss.

We utilize two error metrics to evaluate the prediction models: the average displacement error (ADE) and the final displacement error (FDE). The ADE is computed as the average L2 distance between the ground truth trajectory and the predicted trajectory across all predicted time steps. Conversely, the FDE measures the distance between the predicted final destination and the true final destination at the end of the prediction period $t_{pred}$.

Table~\ref{tab:teacher_preformance} presents the ADE and FDE values for different prediction models over a prediction horizon of $1$ and $6$ time-steps of $0.3$ seconds each. Models with $\mathbb{I}_{adv} = 0$ indicate regression-based training, while models with $\mathbb{I}_{adv} = 1$ indicate adversarial training. We choose the model trained with $k_{var}=1$ and $\mathbb{I}_{adv}=0$ as our main interactive prediction model due to its lowest ADE and FDE on the test data, which is used as a teacher network in the knowledge distillation.

\begin{remark}
Regression-trained models outperform in ADE and FDE due to direct variety loss minimization, while adversarial-trained models may offer advantages like noise robustness and social acceptability of predictions. Thus, model choice should align with specific applications.
\end{remark}
% $\mathbb{E}_{z \sim p(z)}[log(1 - D(G(z)))]$

\begin{table}[]
\centering
% \resizebox{\columnwidth}{!}{%
\footnotesize{
\begin{tabular}{|c|c|cc|cc|}
\hline
\multirow{2}{*}{\textbf{$k_{var}$}} & \multirow{2}{*}{\textbf{$\mathbb{I}_{adv}$}} & \multicolumn{2}{c|}{\textbf{$t_{pred} = 1$}}       & \multicolumn{2}{c|}{\textbf{$t_{pred} = 6$}}       \\ \cline{3-6} 
                                    &                                              & \multicolumn{1}{c|}{\textbf{ADE}}  & \textbf{FDE}  & \multicolumn{1}{c|}{\textbf{ADE}}  & \textbf{FDE}  \\ \hline
1                                   & 1                                            & \multicolumn{1}{c|}{0.02}          & 0.21          & \multicolumn{1}{c|}{0.4}           & 1.41          \\ \hline
\textbf{1}                          & \textbf{0}                                   & \multicolumn{1}{c|}{\textbf{0.01}} & \textbf{0.12} & \multicolumn{1}{c|}{\textbf{0.26}} & \textbf{0.96} \\ \hline
30                                  & 1                                            & \multicolumn{1}{c|}{0.01}          & 0.13          & \multicolumn{1}{c|}{0.27}          & 0.99          \\ \hline
64                                  & 1                                            & \multicolumn{1}{c|}{0.01}          & 0.13          & \multicolumn{1}{c|}{0.26}          & 0.96          \\ \hline
64                                  & 0                                            & \multicolumn{1}{c|}{0.01}          & 0.13          & \multicolumn{1}{c|}{0.26}          & 0.98          \\ \hline
\end{tabular}
}
%
% }
\caption{ADE and FDE of different teacher networks for $t_{pred}=1 \ (0.3 \ sec)$ and  $t_{pred}=6 \ (1.8 \ sec)$.}
\label{tab:teacher_preformance}
\end{table}

\subsection{Student prediction model}\label{subsec:student_network}

\begin{figure}[ht]
    \centering
    \includegraphics[width=0.55\linewidth, height=\linewidth, keepaspectratio]{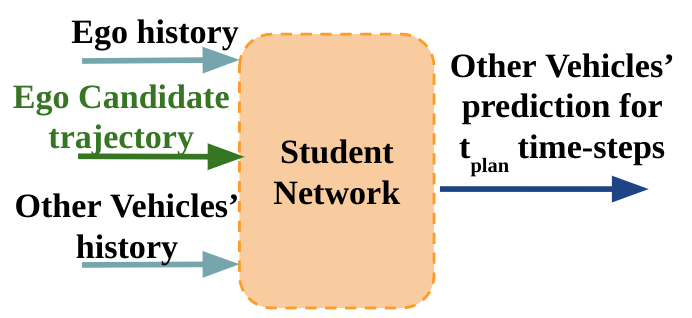}
    \caption{\footnotesize{Student network outputs the interactive trajectory for $t_{plan}$ time-steps by using the ego candidate trajectory as the input. }} 
    \label{fig:student_network}
\end{figure} 

% \begin{figure}
%     \centering
%     \includegraphics[width=1\linewidth, height=\linewidth, keepaspectratio]{images/student_generator.png}
%     \caption{\footnotesize{Overview of the student network interactive trajectory generator. Student network outputs the full interactive trajectory by using the ego candidate trajectory as the input and running the decoder for $t_{plan}$ time steps.}\di{I think it would make sense to put Figure 2 and 4 side by side across the top of the page. However, to really get the impact of the change we might also want to include Fig 1, and a funnel to show all of figure 1 is a stage of figure 2.I'm not sure I explained this well so let me know if you want me to explain more.}} 
%     \label{fig:student_generator}
% \end{figure} 

In order to predict the influence of the ego vehicle's trajectory on other vehicles over a duration of $t_{plan}$ time steps, the teacher network (Fig.~\ref{fig:teacher_network}) employs Social-GAN repeatedly for $t_{plan}$ time steps. This approach significantly amplifies the computational complexity of the prediction inference, as it necessitates running the complete Social-GAN's generator multiple times, as opposed to solely the decoder. Consequently, it is imperative to devise a student network capable of accepting the ego candidate trajectory as input and directly producing the interactive trajectories for all surrounding vehicles as shown in Fig.~\ref{fig:student_network}.

To mitigate the increased computational complexity of the teacher network, we designed a student network with a similar architecture to the teacher network, but with an additional input of the candidate ego trajectory. We made modifications to the decoder of the network, allowing it to internally substitute the output ego prediction each time-step with the corresponding ego position in the candidate trajectory before passing it to the decoder again in the subsequent time step (see Fig.~\ref{fig:student_generator}). This modification in the student network enabled us to generate the complete interaction prediction trajectory by running the decoder for $t_{plan}$ time steps, as opposed to multiple runs of the full Social-GAN's generator.

\textbf{Student Network Training Data:} To obtain training data for the student network, we run the interaction-aware trajectory planning simulation with the teacher network by initializing the scenarios derived from the NGSIM dataset. Using the initial vehicle positions from the scenarios, we randomly assign a target lane for the ego vehicle. The interaction-aware trajectory planner uses Algorithm~\ref{alg:mpc_admm} for motion planning. During the state optimization of the algorithm ($\boldsymbol{\hat{Z}}$ update), different candidate trajectories for the ego vehicle are generated and the teacher network is utilized for the interactive prediction. We record the vehicles' history, ego candidates, and the interactive teacher network predictions, which are used to train the student network. The interactive teacher predictions serve as the ground truth for training. To generate the training data, we establish a fixed manual seed for sampling the latent variable $z$. This ensures the reproducibility and consistency of the teacher network's output, thereby allowing the student network to learn from it.

\begin{figure}[ht]
    \centering
    \includegraphics[width=\linewidth, height=\linewidth, keepaspectratio]{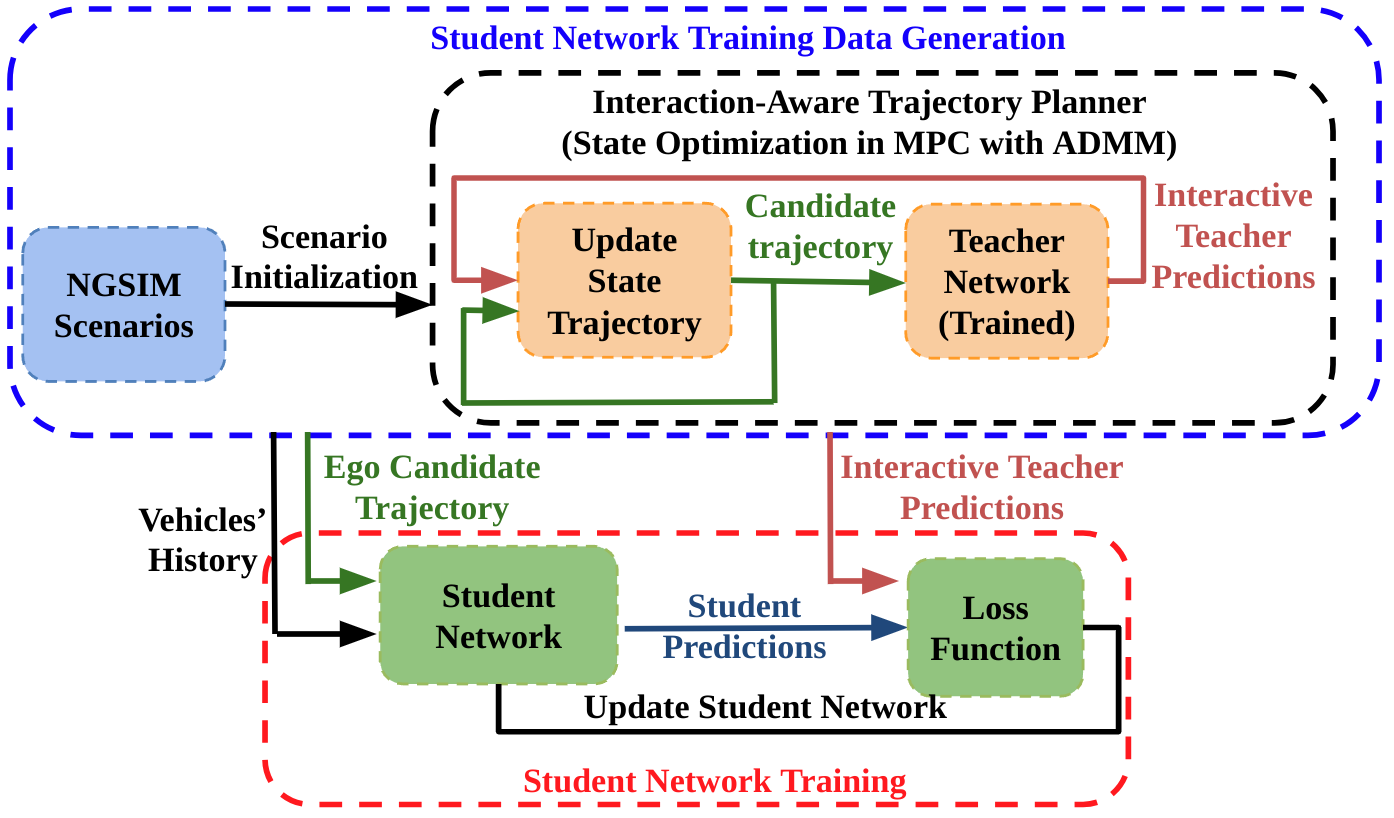}
    \caption{\footnotesize{Training data of the student network is generated by running the MPC with ADMM algorithm with the teacher network using the scenarios initialized from the NGSIM dataset.}} 
    \label{fig:student_data}
\end{figure}

Fig.~\ref{fig:student_data} depicts the process of generating training data for the student network, utilizing the MPC with ADMM algorithm alongside the teacher network. The dataset consists of approximately $568$K training examples, $113$K cross-validation examples, and $116$K test examples.
% The training set consisted of $568005$ examples, the cross-validation set consisted of $112601$ examples, and the test set comprised $115890$ examples. 
% \di{is there a reason for such arbitrary numbers?}

\textbf{Student Network Training:} The student network aims to condense the teacher network; therefore, the hidden state dimensions are reduced to $16$ for the encoder and decoder in the generator, and $24$ for the encoder of the discriminator. The MLP dimension is also decreased to $16$. In order to generate predictions similar to the teacher network, the student network excludes the sampling of the latent variable $z$ and produces a single output prediction ($k=1$). The loss function used is the same as~\eqref{eq:loss}, where the variety loss for $k=1$ simplifies to the L2 loss. Multiple student networks are trained using regression training and adversarial training (GAN) with a discriminator network, mirroring the training approach of the teacher network.

Table~\ref{tab:student_preformance} presents the ADE and FDE values for different prediction models over a prediction horizon of $6$ time-steps of $0.3$ seconds each. We choose the model trained with $k_{var}=128$ and $\mathbb{I}_{adv}=0$ as our main student network due to its lowest ADE and FDE on the test data.

% \usepackage{multirow}
% \usepackage{graphicx}
\begin{table}[]
\centering
% \resizebox{\columnwidth}{!}{%
{\footnotesize
\begin{tabular}{|c|c|cc|}
\hline
\multirow{2}{*}{\textbf{$k_{var}$}} & \multirow{2}{*}{\textbf{$\mathbb{I}_{adv}$}} & \multicolumn{2}{c|}{\textbf{$T_{pred} = 6$}}       \\ \cline{3-4} 
                                    &                                              & \multicolumn{1}{c|}{\textbf{ADE}}  & \textbf{FDE}  \\ \hline
1                                   & 1                                            & \multicolumn{1}{c|}{0.69}          & 1.14          \\ \hline
1                                   & 0                                            & \multicolumn{1}{c|}{0.17}          & 0.31          \\ \hline
64                                  & 1                                            & \multicolumn{1}{c|}{0.24}          & 0.42          \\ \hline
64                                  & 0                                            & \multicolumn{1}{c|}{0.17}          & 0.31          \\ \hline
\textbf{128}                        & \textbf{0}                                   & \multicolumn{1}{c|}{\textbf{0.16}} & \textbf{0.29} \\ \hline
\end{tabular}
}
%
% }
\caption{ADE and FDE of different student networks for $T_{pred}=6 \ (1.8 \ sec)$.}
\label{tab:student_preformance}
\end{table}

\section{Results}\label{sec:results}

\begin{figure}[ht]
    \centering
    \includegraphics[width=0.5\linewidth, height=\linewidth, keepaspectratio]{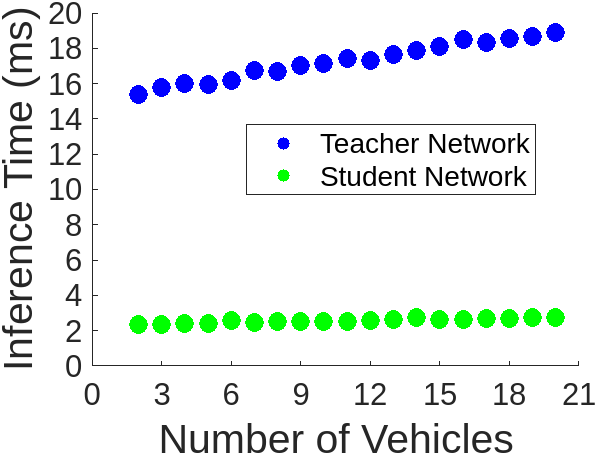}
    \caption{\footnotesize{Inference time comparison between the teacher and the student network for $T_{plan} = 6$ with varying number of vehicles.}} 
    \label{fig:inference_time}
\end{figure}
We evaluate the performance of the interaction-aware trajectory planner using both the teacher and the student network. All evaluations were performed on a single-threaded Intel(R) Xeon(R) CPU E5-2640 v4 @$2.40$ GHz. Fig.~\ref{fig:inference_time} presents a comparison of average interactive prediction inference times between the teacher and the student network for $T_{plan} = 6$, specifically $1.8$ seconds, considering varying numbers of vehicles. The teacher network, due to its repeated utilization of Social-GAN's generator, exhibits an inference time of approximately $15.4-18.9$ ms. In contrast, the student network demonstrates a significantly shorter inference time, approximately $2.3-2.7$ ms.

\begin{figure}[ht]
    \centering
    \includegraphics[width=\linewidth, height=\linewidth, keepaspectratio]{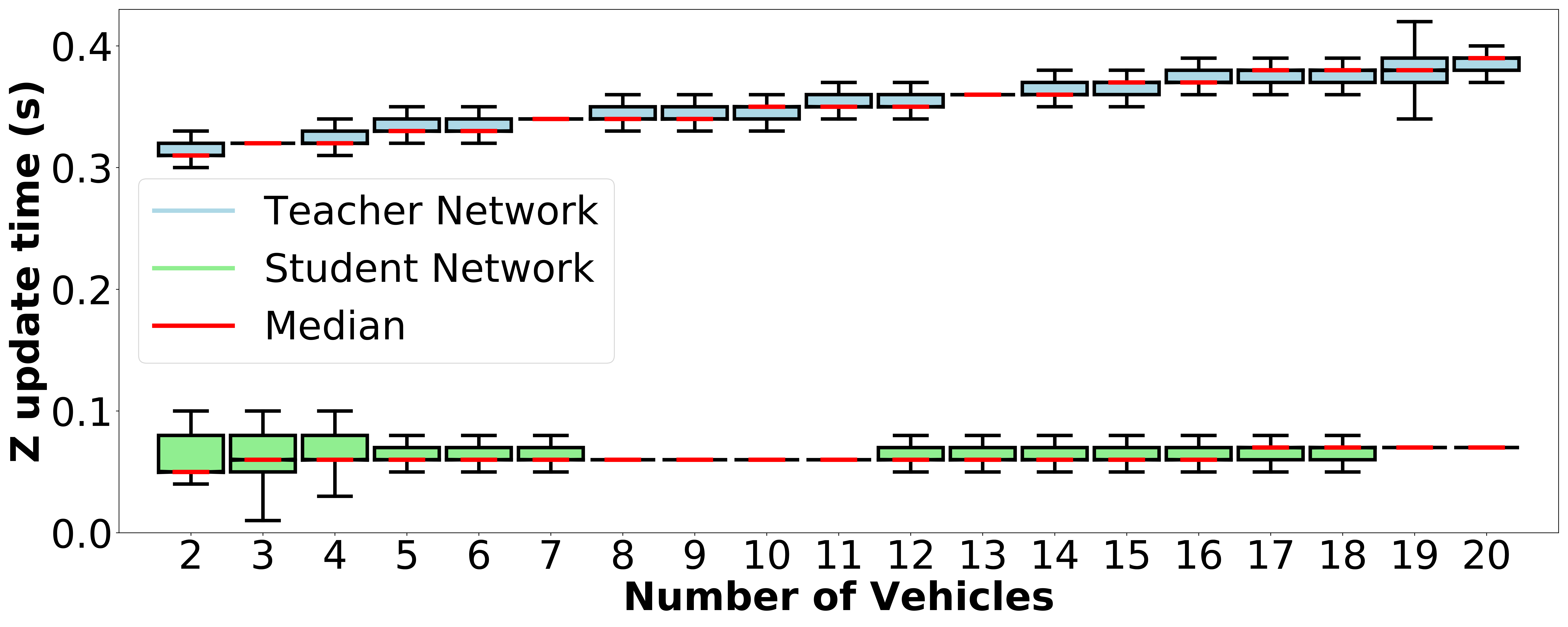}
    \caption{\footnotesize{Z update time (s) for the interaction-aware trajectory planner with the teacher and student network, respectively.}} 
    \label{fig:z_update_time}
\end{figure}

We performed $100$ interaction-aware trajectory planning simulations for each specific number of vehicles, employing both the teacher and student networks. The test scenarios were sourced from the NGSIM dataset and derived from the test dataset of the teacher network. During each simulation, the ego vehicle was assigned a random target lane, and the interaction-aware planner was executed using both the teacher and student network. Each simulation consisted of a range of $9$ to $30$ planning steps, with an average of $16$ steps. Each planning step represented a duration of $0.3$ seconds.

The optimization problem defined by~\eqref{eq:compact_obj}-\eqref{eq:inequality_constraint} was solved using Algorithm~\ref{alg:mpc_admm} for each planning step. During each step, multiple ADMM epochs were executed until convergence, solving the three sub-optimizations discussed in Section~\ref{subsec:interaction_aware_trajectory_planning}. Recall that the state update ($\boldsymbol{\hat{Z}}$) presents a computational bottleneck for the interaction-aware planner due to the inclusion of a neural-network-based prediction model.

To assess the computational efficiency, we gathered data on the state optimization time from approximately $11,000$ state updates for each vehicle in the $100$ simulations conducted using both the teacher and the student network. A box plot visualization in Fig.~\ref{fig:z_update_time} illustrates the distribution of state optimization times, spanning from the first quartile to the third quartile. The red line within each box plot corresponds to the median state update time. Importantly, the student network, benefiting from its faster inference speeds, accelerated the state update optimization process by approximately sixfold in comparison to the teacher network.

\begin{figure}[ht]
    \centering
    \includegraphics[width=\linewidth, height=\linewidth, keepaspectratio]{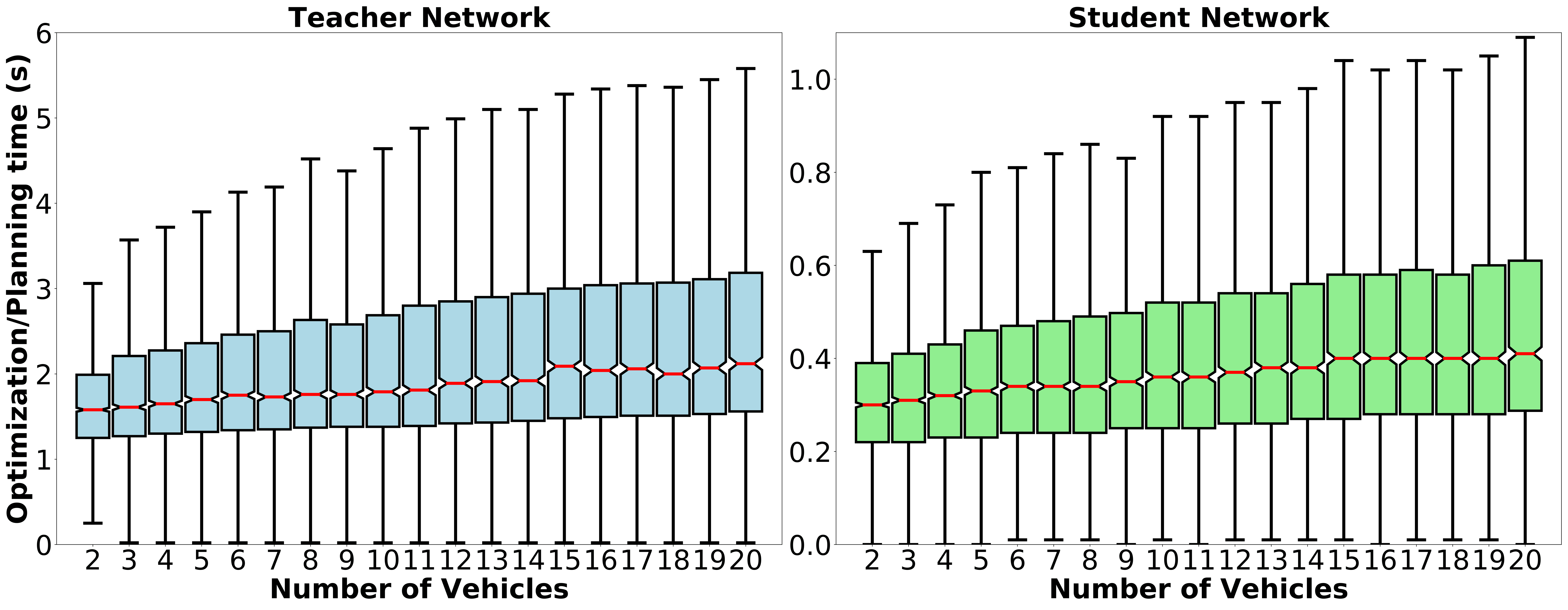}
    \caption{\footnotesize{Optimization time (s) for the interaction-aware trajectory planner with the teacher and student network, respectively.}} 
    \label{fig:optimization_time}
\end{figure}

Figure~\ref{fig:optimization_time} presents a box plot illustrating the distribution of total optimization/planning times. The data was collected from approximately $1,500$ optimizations, performed in $100$ simulations for each vehicle using both the teacher (left) and the student network (right). Notably, the interaction-aware planner utilizing the student network achieves a higher frequency of operation, running at approximately $2-4$ Hz. In contrast, the planner with the teacher network operates at a lower frequency of around $0.5-0.7$ Hz, demonstrating an improvement of approximately fivefold. 

It is important to note that in our simulations, we utilize an initial solution guess for the optimizer obtained by extrapolating the ego vehicle's trajectory under the assumption of constant velocity. Consequently, the initial planning steps of the simulation require more time to compute the solution compared to the subsequent steps. To further enhance optimization time, providing a better initial guess using a faster heuristic algorithm could be employed.

\begin{figure}[ht]
    \centering
    \includegraphics[width=\linewidth, height=\linewidth, keepaspectratio]{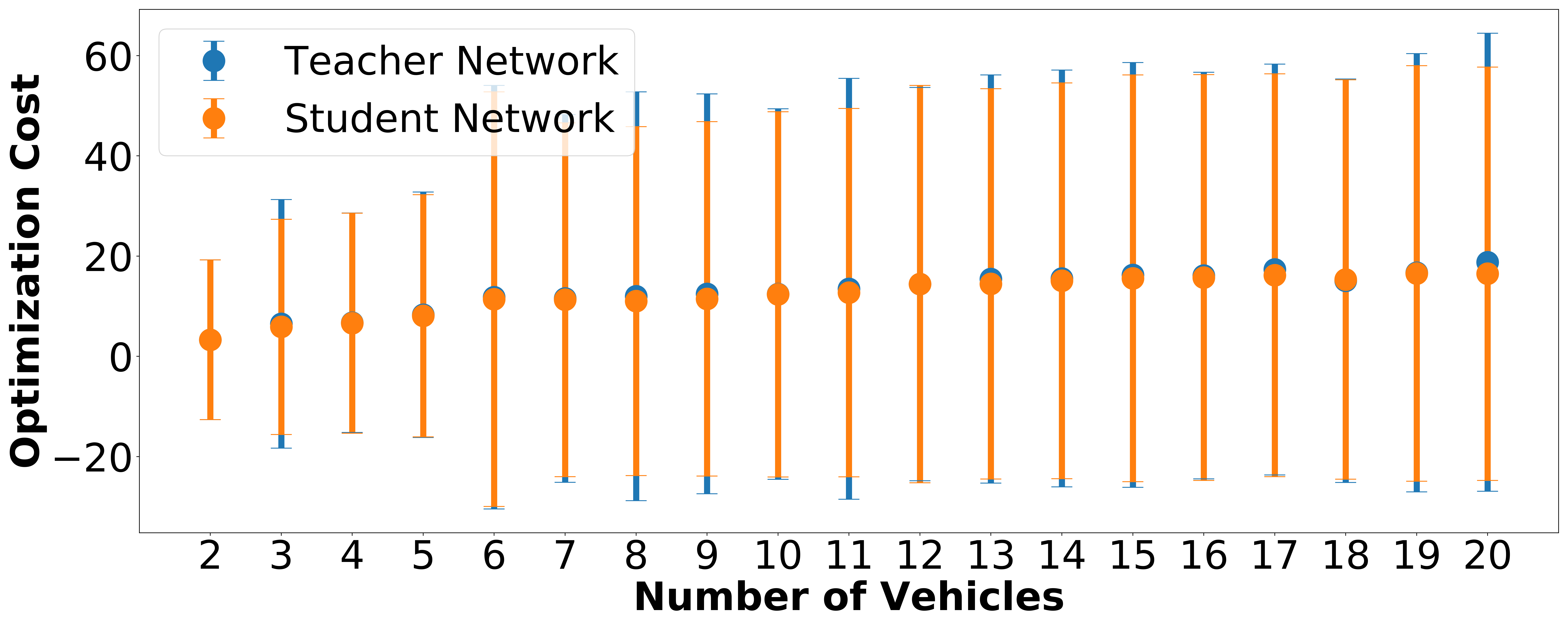}
    \caption{\footnotesize{Optimization cost for the interaction-aware trajectory planner with the teacher and student network, respectively.}} 
    \label{fig:optimization_cost}
\end{figure}

Lastly, Fig.~\ref{fig:optimization_cost} displays the mean and standard deviations of the optimal cost for the planning steps using both the teacher and the student network. It is evident that both networks produce optimization solutions with similar costs, indicating comparable optimal trajectories. Consequently, the use of the smaller student network, trained through knowledge distillation from the teacher network, significantly accelerates the optimization process without any significant loss in accuracy.

\section{Conclusion}\label{sec:conclusions}

In many engineering applications, the integration of neural-network-based systems with traditional optimization frameworks often gives rise to constrained optimization problems involving neural networks. These problems are inherently challenging to solve in real-time due to the complexity and non-convex nature of neural networks. Our focus lies specifically on one such problem, namely interaction-aware trajectory planning in autonomous driving.

In this study, we investigated the impact of knowledge distillation on constrained optimization by training a smaller student network. The student network leverages knowledge acquired from a larger prediction model known as the teacher network, and its purpose is to generate interactive predictions for a given history of surrounding vehicles and the planned trajectory of the autonomous vehicle over a planning horizon. Our findings reveal that by carefully designing the student network, we can significantly reduce the inference time compared to the teacher network. This acceleration greatly enhances the efficiency and scalability of the interaction-aware trajectory planner, enabling its real-time implementation in practical applications.

\bibliographystyle{ieeetr}
\bibliography{aaai23.bib}

\end{document}